\documentclass{article} 
\usepackage[preprint]{colm2026_conference}

\usepackage{microtype}
\usepackage{hyperref}
\usepackage{url}
\usepackage{booktabs}
\usepackage{amsmath} 
\usepackage{pifont}
\usepackage{algorithm}
\usepackage{algorithmic}
\usepackage{graphicx}
\usepackage{wrapfig}
\usepackage{multirow}
\usepackage{booktabs}
\usepackage{arydshln}

\usepackage{cleveref}

\usepackage{caption}
\newcommand{\mask}{\texttt{M}}
\newcommand{\cM}{\mathcal{M}}
\newcommand{\cV}{\mathcal{V}}

\newcommand{\cD}{\mathcal{D}}

\usepackage{lineno}

\definecolor{darkblue}{rgb}{0, 0, 0.5}
\hypersetup{colorlinks=true, citecolor=darkblue, linkcolor=darkblue, urlcolor=darkblue}

\title{BackPlay: Head-Only Look-Back Self-Correction for \\Diffusion Language Models\thanks{Work done at Amazon. Correspondence to \texttt{lliu606@amazon.com,\texttt{lliu606@gatech.edu}}.}}

\author{Liming Liu, Binxuan Huang, Zixuan Zhang, Xin Liu, Bing Yin, Tuo Zhao}

\begin{document}

\ifcolmsubmission
\linenumbers
\fi

\maketitle

\begin{abstract}
Diffusion Language Models (DLMs) decode multiple tokens in parallel, but aggressive multi-token decoding amplifies cross-token dependency errors and can sharply degrade generation quality. We propose BackPlay, a frozen-backbone self-correction framework that trains only a lightweight correction head on a finetuned DLM without updating any backbone or adapter parameters. Because the head is trained on errors produced by the same frozen generator used at inference time, its training distribution aligns with the error patterns of the deployed model. We further introduce Look-back Correction, a training mechanism that injects predictions from earlier, more corrupted denoising states into later, richer contexts, enabling the head to leverage later context to detect mistakes made in earlier generation steps. During inference, BackPlay periodically revisits previously generated tokens through selective remasking and regeneration to limit error accumulation. Across mathematical reasoning and code generation benchmarks, BackPlay improves the speed--quality trade-off of the underlying DLM under  multi-token decoding.
\end{abstract}

\section{Introduction}

Diffusion Language Models (DLMs) have emerged as a powerful alternative to autoregressive (AR) models for language generation, offering substantial speedups through parallel decoding \citep{sahoo2024simple, nie2024scaling, ye2025dream,zhu2025llada}. Unlike AR models that generate tokens sequentially, DLMs start from a fully masked sequence and iteratively reveal tokens through a denoising process, producing multiple tokens simultaneously at each step. This parallel generation enables faster inference while achieving competitive performance across reasoning, coding, and planning tasks \citep{nie2025large, ye2025dream, zhu2025llada}.

The efficiency gains of parallel generation come with inherent challenges. When DLMs generate multiple tokens in the same iteration, particularly under aggressive multi-token decoding, they can amplify cross-token dependency errors \citep{liu2024discrete, xu2024energy, kang2025parallelbench}. Errors introduced in early denoising steps may persist and propagate through subsequent iterations, degrading generation quality. This motivates self-correction mechanisms that can detect and rectify erroneous predictions during inference. In particular, tokens that appear locally plausible under sparse early context may become globally inconsistent as later denoising steps reveal richer surrounding information.

Self-correction methods for DLMs span a design space defined by several choices. Training-free approaches refine outputs via heuristic remasking or model-derived confidence scores \citep{wang2025remasking, zhao2024informed, peng2025path}, but do not explicitly train a token-correctness predictor. Trainable approaches add a correction head that learns token-level error detection, with some methods jointly updating the generator alongside the head via a multi-objective loss \citep{huang2025don, kim2025fine}. These methods differ along several design axes: whether the generator parameters are updated during corrector training, how artifact-containing training sequences are constructed, and whether the corrector explicitly leverages later denoising context to judge earlier predictions.

This paper targets a specific setting: a post-trained DLM is already available, and we wish to add inference-time self-correction without modifying the deployed generator. This motivates two design desiderata: (i)~the corrector should train on a fixed generator to avoid distribution drift during head training, and (ii)~the corrector should leverage later denoising context to revisit earlier predictions. In the head-based setup studied here, we empirically observe a fidelity--correction trade-off under joint optimization on the 1B model (\Cref{tab:generative_fidelity}), which means that simultaneously optimizing generation and correction objectives can degrade DLM’s generative capability. Moreover, preserving the deployed generator can be operationally preferable when the model has undergone substantial post-training.

\textbf{Contributions:} We propose BackPlay, a head-only self-correction framework for frozen finetuned DLMs. Our contributions are threefold:

First, we decouple corrector training from the generator. By freezing the finetuned DLM and training only a lightweight correction head---without updating backbone or adapter parameters---we avoid generator drift during head training and remove generator-side gradient computation by construction.

Second, we introduce Look-back Correction, a training mechanism that injects predictions from earlier, more corrupted denoising states into later, richer contexts, providing explicit later-context supervision for earlier errors. This enables the head to detect mistakes plausible under sparse context but inconsistent under richer later context.

Third, we show that BackPlay improves the speed--quality trade-off over the SFT baseline and internal ablations on mathematical reasoning and code generation benchmarks.

\looseness=-1 BackPlay is not the first trainable self-correction method for DLMs. Relative to PRISM \citep{kim2025fine} and RemeDi \citep{huang2025don}, our focus is a distinct operating regime: a frozen deployed generator, head-only training, and explicit later-context supervision for earlier errors. We summarize these design differences in \Cref{tab:method_comparison} and detail them in \Cref{sec:DSC,sec:FCA}. Accordingly, this paper studies whether inference-time self-correction can be added to a post-trained DLM without updating the generator, while allowing the corrector to revisit earlier decisions under later context.

\section{Preliminaries and Background on Trainable Self-Correction}
\label{relatedwork}

\subsection{Diffusion Language Models (DLMs)}

Diffusion Language Models (DLMs) \citep{nie2024scaling, nie2025large, ye2025dream, zhu2025llada} extend discrete diffusion \citep{austin2021structured, lou2023discrete, he2023diffusionbert, sahoo2024simple, gat2024discrete, shi2024simplified} to language generation. A forward corruption process independently replaces tokens with a special mask token $\mask$, and the model is trained to recover the masked tokens from the remaining context.

\textbf{Forward process.} Let $\mathbf{x}_0 \in \cV^L$ be a length-$L$ clean sequence from dataset $\cD$ with vocabulary $\cV$. For time $t\in[0,1]$, the corrupted sequence $\mathbf{x}_t \in (\cV\cup\{\mask\})^L$ is obtained by independently masking each position:
\begin{align*}
q_{t|0}(\mathbf{x}_t | \mathbf{x}_0) = \prod_{i=1}^L q_{t|0}(x_t^i | x_0^i), \quad\quad
q_{t|0}(x_t^i | x_0^i) =
\begin{cases}
\alpha_t, & x_t^i=x_0^i,\\
1-\alpha_t, & x_t^i= \mask.
\end{cases}
\end{align*}
Following standard practice, the linear schedule $\alpha_t = 1-t$ is adopted, so that $t=1$ yields a fully masked sequence.

\textbf{Training.} The model $p_\theta(\mathbf{x}_0| \mathbf{x}_t)$ is trained to predict original tokens at masked positions by minimizing the demasking loss:
\begin{equation}
\label{eq:sft loss}
\begin{aligned}
\mathcal{L}_{\text{Demask}}(\theta)
= \mathbb{E}\Big[
\frac{1}{t}
\sum_{i:\, x_t^i=\mask}
-\log p_\theta(x_0^i \mid \mathbf{x}_t)
\Big],
\end{aligned}
\end{equation}
where $t \sim \mathrm{Unif}[0,1]$, $\mathbf{x}_0 \sim \mathcal{D}$, and $\mathbf{x}_t \sim q_{t|0}(\cdot|\mathbf{x}_0)$.

\textbf{Inference.} Generation starts from a fully masked sequence $\mathbf{x}_1 = (\mask, \dots, \mask)$ and proceeds over decreasing time steps $t_0{=}1 > \cdots > t_N{=}0$. At each step, a subset of masked positions is selected and their tokens are sampled from $p_\theta(x_0^i \mid \mathbf{x}_{t_\ell})$. We refer to positions with $x_t^i \neq \mask$ as \emph{visible} and positions with $x_t^i = \mask$ as \emph{masked} in the following sections.


Because DLMs update multiple positions in parallel from the same partially observed state $\mathbf{x}_t$, the resulting tokens may violate cross-token dependencies even when each per-position prediction is locally plausible \citep{liu2024discrete, xu2024energy, kang2025parallelbench}. This motivates correction mechanisms that evaluate visible tokens during denoising.

\subsection{Self-Correction of DLMs}

Training-free self-correction methods refine DLM outputs via heuristic remasking or model-derived confidence scores \citep{zhao2024informed, wang2025remasking, peng2025path}, but are not directly trained against token correctness, which can limit calibration and systematic improvement. Trainable methods instead augment the DLM with a correction module that learns to predict per-token correctness from data \citep{huang2025don, kim2025fine}.

Concretely, a correction module $\mathbf{g}_{\theta,\phi}(\mathbf{z}_t)$ takes a partially denoised sequence $\mathbf{z}_t$ and, for each visible position $i$ (i.e., $z_t^i \neq \mask$), outputs a \emph{correctness score}
\[
c_i \;=\; \mathbf{g}_{\theta,\phi}(\mathbf{z}_t)_i \;\in\; [0,1],
\]
interpreted as $c_i \approx P(z_t^i = x_0^i \mid \mathbf{z}_t)$. At inference time, an equivalent \emph{error score} $e_i := 1 - c_i$ can be used to rank candidates for remasking. The module is trained using Binary Cross-Entropy (BCE) loss over visible positions:
\begin{equation}
\label{eq:loss error}
    \mathcal{L}_{\text{Err}}(\theta, \phi) = \mathbb{E} \Big[ \sum_{i:\,z^i_{t} \neq \mask} \text{BCE}\!\left(c_i,\; \mathbb{I}[z^i_{t} = x^i_{0}]\right) \Big],
\end{equation}
where $t \sim \mathrm{Unif}[0,1]$, $\mathbf{x}_0 \sim \mathcal{D}$, and $\mathbf{z}_t \sim q_{\mathrm{sc}}(\cdot \mid \mathbf{x}_0, t)$. Masked positions ($z_t^i = \mask$) contribute no loss, as there is no visible token to evaluate. Throughout this paper, correctness during training refers to exact agreement with the reference token $x_0^i$.

The central design problem is the choice of the self-correction training distribution $q_{\mathrm{sc}}(\mathbf{z}_t \mid \mathbf{x}_0, t)$, which specifies how artifact-containing sequences $\mathbf{z}_t$ are constructed. Here, $q_{\mathrm{sc}}$ denotes the distribution over artifact-containing sequences $\mathbf{z}_t$ induced by the full construction pipeline, including artifact generation, position selection, and insertion. Prior methods start with a corrupted sequence $\mathbf{x}_t \sim q_{t|0}(\cdot | \mathbf{x}_0)$ and introduce artifact tokens $\mathbf{y} \sim P_{\rm artifact}$ at selected positions $\cM$:
\begin{align*}
    \mathbf{z}_t &= \text{replace}(\mathbf{x}_{t},\; \mathbf{y},\; \mathcal{M}), \quad \text{where}\quad[\text{replace}(\mathbf{x}_{t}, \mathbf{y}, \mathcal{M})]_i = \begin{cases}
        y^i, & \text{if } i \in \mathcal{M},\\
        x_t^i, & \text{otherwise.}
    \end{cases}
\end{align*}
RemeDi \citep{huang2025don} draws artifacts from a uniform distribution over the vocabulary $\cV$, providing a simple control distribution that does not explicitly model structured, model-generated errors. PRISM \citep{kim2025fine} instead generates artifacts from the model's own predictions at a slightly more corrupted state: given a step size $\Delta t$,
$$\mathbf{x}_{t+\Delta t} \sim q_{t+\Delta t|0, t}(\cdot|\mathbf{x}_0, \mathbf{x}_t), \quad \mathbf{y} \sim  p_\theta(\mathbf{x}_0 | \mathbf{x}_{t+\Delta t}),$$
and selects $\cM$ from newly unmasked positions---those masked in $\mathbf{x}_{t+\Delta t}$ but visible in $\mathbf{x}_{t}$---with $|\cM| = \lceil L \cdot \Delta t \rceil$ to simulate one reverse-process transition. Both RemeDi and PRISM jointly optimize the generator and correction head:
\begin{equation}
    \label{eq:total loss}
    \mathcal{L}(\theta, \phi) = \mathcal{L}_{\text{Demask}}(\theta) + \gamma \, \mathcal{L}_{\text{Err}}(\theta, \phi),
\end{equation}
where $\gamma$ balances the demasking and error-detection objectives.

\looseness=-1 These methods differ along three main axes: (a)~whether generator-side parameters $\theta$ are updated during corrector training, (b)~how the self-correction training distribution $q_{\mathrm{sc}}$ is constructed, and (c)~whether earlier predictions are evaluated under later denoising context. These axes motivate the design considerations discussed next.

\subsection{Design Motivation for Frozen-Backbone Self-Correction}
\label{sec:limitations}

Rather than claiming universal limitations of prior methods, we highlight three design considerations that motivate the operating point studied in this paper. In the setting considered here---a post-trained DLM that should remain unmodified---these considerations lead to a frozen-backbone, head-only correction design with explicit later-context supervision.

\textbf{Preserving a deployed post-trained generator.}
When a DLM has undergone substantial supervised finetuning or post-training, it can be desirable to keep the generator parameters $\theta$ fixed and train only the correction head $\phi$. In the head-based setup studied here, \Cref{tab:generative_fidelity} in Section~4.1 provides evidence that joint optimization of $\theta$ and $\phi$ via \Cref{eq:total loss} can introduce a fidelity--correction trade-off on the 1B model.

\textbf{Training the head on a fixed error source.}
When generator-side parameters are updated during head training, the artifact distribution seen by the head can evolve over the course of training. A frozen-generator design removes this source of drift by construction, so that the head trains on errors produced by the same fixed generator used at deployment.

\textbf{Evaluating earlier decisions under later denoising context.}
Some early token decisions become identifiable as errors only after more context has been revealed in later denoising steps. Here, ``look-back'' refers to revisiting earlier denoising decisions under later denoising context; it is defined over the denoising timeline and is orthogonal to the semi-autoregressive block length used at inference. This motivates training recipes that reassess earlier predictions using richer later context.

These considerations define the operating point of BackPlay relative to prior trainable correctors. Our empirical evaluation in Section~4 focuses on controlled baselines and ablations within the frozen-backbone setting.

\section{Method}

\looseness=-1 Section~\ref{sec:DSC} defines a head-only correction objective on a frozen deployed DLM. Section~\ref{sec:FCA} constructs the Look-back Correction (LBC) training distribution that provides later-context supervision for earlier errors. Section~\ref{sec:inference} converts the learned correctness scores into a conservative remasking policy at inference time.

\subsection{Head-Only Self-Correction on a Frozen Deployed DLM}
\label{sec:DSC}

Motivated by the design considerations in \Cref{sec:limitations}---preserving a deployed generator and training the head on a fixed error source---BackPlay treats the finetuned DLM as a frozen feature backbone and appends an auxiliary correction head $\phi$, trained without any backbone or adapter modification, to predict the correctness of currently visible tokens.

Given a finetuned DLM with frozen parameters $\theta^*$, we introduce a shallow Transformer correction head $\phi$ that operates on the penultimate-layer hidden states. Let $N_L$ denote the number of layers in the frozen DLM. The correction head is defined as:
\begin{equation}
\label{eq: head}
\mathbf{g}_{\theta^{*}, \phi}(\mathbf{z}_t) = \sigma\!\left(\phi(\mathbf{h}^{N_L-1})\right) \in [0,1]^L,
\end{equation}
where $\sigma$ denotes the sigmoid function applied position-wise, and $\mathbf{h}^{N_L-1}$ are the hidden states from the forward pass of $p_{\theta^*}(\mathbf{x}_{0} | \mathbf{z}_{t})$. Since $\theta^*$ is fully frozen, the hidden-state distribution seen by $\phi$ does not change during head training, avoiding generator drift.

For each visible position $i$ in the set $\mathcal{V}_t := \{j : z_t^j \neq \mask\}$, the head outputs a correctness score
$c_i = \mathbf{g}_{\theta^{*},\phi}(\mathbf{z}_t)_i \in [0,1]$,
interpreted as $c_i \approx P(z_t^i = x_0^i \mid \mathbf{z}_t)$. The head $\phi$ is optimized using a Binary Cross-Entropy (BCE) loss over visible positions:
\begin{align}
    \label{eq: Dsc loss error}
    \mathcal{L}(\phi) = \mathbb{E} \Big[ \sum_{i \in \mathcal{V}_t} \text{BCE}(c_i, \;\mathbb{I}[z^i_{t} = x^i_{0}]) \Big]
\end{align}
where $t \sim \mathcal{U}[0, 1{-}\Delta t]$, $\mathbf{x}_0 \sim \mathcal{D}$, and $\mathbf{z}_t \sim q_{\mathrm{LBC}}(\cdot \mid \mathbf{x}_0, t;\, \theta^*, \Delta t)$ is the LBC training distribution defined in Section~\ref{sec:FCA}. Masked positions ($z_t^i = \mask$) contribute no loss for $\phi$.

At inference time, correctness scores are converted to error scores via $e_i = 1 - c_i$, and a small subset of visible tokens with the largest $e_i$ are selectively remasked (Section~\ref{sec:inference}).

This design has three structural consequences. First, because $\theta^*$ is frozen, the generative ability of $\theta^*$ is preserved. Second, the head trains on errors induced by the same fixed generator used at deployment, with no generator drift during head training. Third, training only the head eliminates gradient computations in the backbone, thereby accelerating the training process by reducing both computational load and activation memory usage.

\subsection{Look-back Correction}
\label{sec:FCA}

We now define the Look-back Correction (LBC) training distribution $q_{\mathrm{LBC}}$ referenced in \Cref{eq: Dsc loss error}, which addresses the third design consideration in \Cref{sec:limitations}: evaluating earlier decisions under later denoising context. Because the reverse denoising process runs from $t=1$ (fully masked) to $t=0$ (fully revealed), a less corrupted state $\mathbf{x}_t$ corresponds to a later denoising stage with richer context than a more corrupted state $\mathbf{x}_{t+t'}$.

Recall that $\theta^{*}$ denotes the frozen DLM parameters and $\Delta t$ is the step size. For each training sample, we draw $t \sim \mathcal{U}[0, 1{-}\Delta t]$ and $\mathbf{x}_t \sim q_{t|0}(\cdot|\mathbf{x}_0)$. We then sample a forward interval $t' \sim \mathcal{U}[\Delta t, 1{-}t]$ to obtain a more corrupted state $\mathbf{x}_{t+t'} \sim q_{t+t'|0, t}(\cdot|\mathbf{x}_0, \mathbf{x}_{t})$, and generate artifact tokens using the frozen DLM:
\begin{align*}
 \mathbf{y} \sim p_{\theta^*}(\mathbf{x}_{0} | \mathbf{x}_{t+t'}).
\end{align*}
We define the set of \emph{newly unmasked positions}---those masked in $\mathbf{x}_{t+t'}$ but visible in $\mathbf{x}_{t}$:
\[
\mathcal{I}_{\text{new}} = \{i : x_{t+t'}^i = \mask \;\text{and}\; x_t^i \neq \mask\}.
\]
We select $\mathcal{M} \subseteq \mathcal{I}_{\text{new}}$ with $|\mathcal{M}| = \min(\lceil L \cdot \Delta t \rceil,\, |\mathcal{I}_{\text{new}}|)$ by choosing positions where the frozen model exhibits the highest prediction confidence:
\[
\mathrm{conf}_i = \max_{v \in \cV}\, p_{\theta^*}(v \mid \mathbf{x}_{t+t'}), \quad i \in \mathcal{I}_{\text{new}},
\]
following the confidence-based strategy of \citet{nie2025large}. The training sequence $\mathbf{z}_t = \text{replace}(\mathbf{x}_{t},\, \mathbf{y},\, \mathcal{M})$ is formalized in \Cref{alg:lbc_training}.

\looseness=-1 The resulting $\mathbf{z}_t$ is a training-time proxy for a common inference scenario: tokens predicted from a more uncertain state $\mathbf{x}_{t+t'}$ persist into the later, richer context of $\mathbf{x}_t$. Ground-truth context in $\mathbf{x}_t$ outside $\mathcal{M}$ is retained so that the correction head $\phi$ is supervised to detect earlier mistakes by cross-referencing them with surrounding later context. 


Unlike one-step artifact construction, LBC explicitly inserts predictions from a more corrupted state into a later, cleaner state, so the head is supervised to detect earlier mistakes under richer later context. We summarize the design differences between BackPlay and prior trainable self-correction methods in \Cref{tab:method_comparison}.

\begin{table}[htb!]
\centering
\caption{Design comparison of trainable self-correction methods for DLMs.}
\label{tab:method_comparison}
\resizebox{\linewidth}{!}{
\begin{tabular}{l|ccc}
\toprule
 & RemeDi & PRISM & BackPlay (Ours) \\
\midrule
Trainable parameters
& Full model ($\theta + \phi$)
& $\phi$ + LoRA adapters
& $\phi$ only (frozen $\theta$) \\
Generator updated during head training
& Yes (full)
& Yes (LoRA)
& No \\
Artifact source
& Uniform noise
& Training-time generator
& Frozen deployed generator \\
Later-context supervision
& No
& No
& Yes \\
\bottomrule
\end{tabular}}
\end{table}

\subsection{Inference with Conservative Adaptive Remasking}\label{sec:inference}

\looseness=-1 We deploy $(\theta^*,\phi^*)$ with an inference procedure that interleaves generation and correction. At each denoising step, both the generator output $p_{\theta^*}(\mathbf{x}_0 \mid \mathbf{x})$ and the correctness scores $\mathbf{c} = \mathbf{g}_{\theta^*,\phi^*}(\mathbf{x})$ are computed from the same forward pass on the current state $\mathbf{x}$, before any new tokens are revealed. For each visible position $i \in \mathcal{V} = \{j : x^{(j)} \neq \mask\}$, we convert correctness scores to error scores via $e_i = 1 - c_i$. A remasking budget $K$ and error threshold $\tau$ then select the top-$K$ visible tokens with the highest $e_i$, subject to $e_i > \tau$, for remasking. Newly generated tokens in the current step are not evaluated until a later correction round, after more surrounding context has accumulated.

After remasking, the denoising timeline accounts for both the forward step and any rollback from remasking:
\begin{equation}
    t_{\text{next}} = \max\{t - \Delta t + \frac{|\mathcal{M}_{\text{remask}}|}{L}, \, 0\},
\end{equation}
where $L$ is the generation length. This allows the model to revisit remasked positions in subsequent steps with richer context. To avoid intervening when context is still sparse, correction is triggered only every $d$ steps (when $N > 0$ and $N \bmod d = 0$).

\looseness=-1 The four policy hyperparameters serve distinct roles. The threshold $\tau$ controls the precision of error detection: only tokens with $e_i > \tau$ are remasked, preventing over-correction. The budget $K$ limits the number of tokens remasked per correction round, bounding each rollback. The stride $d$ ensures that sufficient new context accumulates between successive correction rounds. The block buffer $\mathcal{H}_{\text{block}}$ (capacity $B$) suppresses oscillation, as discussed below. We keep $(\tau, K, d, B)$ fixed across tasks and tokens-per-step settings; varying $\tau$ from 0.75 to 0.6 or varying $B$ from 4 to 6 on MBPP yields a slight trade-off shift, not brittle behavior (Appendix).

\textbf{Anti-Repetitive Correction.} Oscillation can arise at ambiguous or weakly determined positions where the head repeatedly assigns low correctness while the generator regenerates the same token, because the surrounding context has not changed enough to alter either prediction. To prevent repeated rollback on unchanged evidence, we maintain a buffer $\mathcal{H}_{\text{block}}$ of recently remasked indices (capacity $B$) and temporarily exclude them from remasking candidates. This ensures the remasking budget is spent on distinct positions.

We summarize the full inference procedure in \Cref{alg:dsc_inference}.

\begin{algorithm}
\caption{Inference with Conservative Adaptive Remasking}
\label{alg:dsc_inference}
\begin{algorithmic}[1]

\REQUIRE
Frozen DLM $\theta^*$,
correction head $\phi^*$,
tokens per step $k$,
generation length $L$,
remasking budget $K$,
error threshold $\tau$,
correction stride $d$,
block-buffer capacity $B$.

\STATE $\Delta t \leftarrow k/L$, \quad $t \leftarrow 1$, \quad $N \leftarrow 0$, \quad $\mathbf{x} \leftarrow [\text{MASK}]^L$, \quad $\mathcal{H}_{\text{block}} \leftarrow \emptyset$

\WHILE{$t > 0$}
    \STATE $\mathcal{V} \leftarrow \{i : x^{(i)} \neq [\text{MASK}]\}$; \; $\mathcal{M} \leftarrow \{i : x^{(i)} = [\text{MASK}]\}$; \; $\mathcal{M}_{\text{remask}} \leftarrow \emptyset$
    \STATE $P_{\text{demask}} \leftarrow p_{\theta^*}(\mathbf{x}_{0} \mid \mathbf{x})$; \; $\mathbf{c} \leftarrow \mathbf{g}_{\theta^*,\phi^*}(\mathbf{x})$ \COMMENT{shared forward pass}
    \STATE $\mathcal{I} \leftarrow \text{arg-top-}k_{i \in \mathcal{M}} \!\left(\max_v P_{\text{demask}}^{(i)}(v) \right)$; \; sample $x^{(i)} \sim P_{\text{demask}}^{(i)}$ for $i \in \mathcal{I}$
    \IF{$N > 0$ \textbf{and} $N \bmod d = 0$}
        \STATE $\mathbf{e} \leftarrow 1 - \mathbf{c}$; \; $\mathcal{C} \leftarrow \mathcal{V} \setminus \mathcal{H}_{\text{block}}$ \COMMENT{error scores; eligible candidates}
        \STATE $\mathcal{M}_{\text{candid}} \leftarrow \text{arg-top-}\min(K, |\mathcal{C}|)_{i \in \mathcal{C}}(e_i)$; \; $\mathcal{M}_{\text{remask}} \leftarrow \{i \in \mathcal{M}_{\text{candid}} : e_i > \tau\}$
        \FOR{$i \in \mathcal{M}_{\text{remask}}$}
            \STATE $x^{(i)} \leftarrow [\text{MASK}]$; \; append $i$ to $\mathcal{H}_{\text{block}}$; \; if $|\mathcal{H}_{\text{block}}| \ge B$ remove oldest
        \ENDFOR
    \ENDIF
    \STATE $t \leftarrow \max\{t - \Delta t,\, 0\} + |\mathcal{M}_{\text{remask}}|/L$; \; $N \leftarrow N + 1$
\ENDWHILE

\end{algorithmic}
\end{algorithm}

\section{Experiments}

\looseness=-1 We evaluate BackPlay in three complementary ways: (i)~a controlled frozen-vs-joint training ablation on SMDM 1B (\Cref{sec:SMDM}), (ii)~end-to-end accuracy--efficiency comparisons against SFT baselines on LLaDA 8B for coding and mathematical reasoning (\Cref{sec:LLaDA}), and (iii)~controlled ablations on artifact construction (\Cref{sec:ablation}) and remasking policy (\Cref{sec:random}). We do not include direct reproductions of PRISM or RemeDi under matched settings; accordingly, all empirical claims are relative to the reported baselines and controls.

\looseness=-1 Throughout Section~4, we report zero-shot accuracy (Pass@1) and $\text{Iter}_{\text{avg}}$, the average number of DLM backbone forward passes. The correction head reuses the penultimate hidden states from the same forward pass and does not incur an additional backbone forward pass. In our LLaDA experiments, we follow the standard semi-autoregressive decoding protocol with block length 32 and maximum generation length 1024. BackPlay's ``look-back'' operates over denoising-time context within this protocol, not the block length.

\subsection{SMDM 1B: Controlled Ablation of Frozen vs.\ Joint Training}
\label{sec:SMDM}

This experiment serves as a controlled ablation of optimization strategy. Following the setting in \citet{nie2024scaling}, we adopt the 1B pretrained SMDM model as the base and perform standard SFT on augmented data \citep{deng2023implicit} to obtain the finetuned model $\theta^*$. Based on $\theta^*$, we apply our training recipe in \Cref{sec:DSC} and \Cref{sec:FCA} to learn the correction head $\phi^*$. As a control, we study joint optimization by training $\theta$ and $\phi$ simultaneously with the loss in \Cref{eq:total loss}, denoting the final model as $(\theta + \phi)^*$.

\looseness=-1 BackPlay and the joint-training control share the same base model, training data, LBC artifact construction, correction-head architecture, and inference policy; they differ only in whether the DLM parameters $\theta$ are updated during correction-head training. We do not claim that joint optimization is universally harmful; rather, this experiment isolates the frozen-vs-joint training effect in our head-based SMDM 1B setup.

Following \Cref{sec:DSC}, the input to $\phi$ is the hidden state from the 19th transformer block (out of 20) of the SMDM 1B model. The correction head $\phi$ is a 2-layer transformer. For both BackPlay and the joint-training control, training sequences $\mathbf{z}_t$ are generated using the LBC method in \Cref{sec:FCA}. During inference, without self-correction, all models use the confidence-based strategy in \Cref{alg:confidence_inference} \citep{nie2025large}. With self-correction, all models adopt the conservative adaptive remasking procedure in \Cref{alg:dsc_inference}. Hyperparameters are provided in \Cref{sec:infer smdm}.

\subsubsection{Frozen vs.\ Joint Training on GSM8K}
\label{sec: 1B result}

\begin{table}[htb!]
    \caption{Controlled ablation on GSM8K in our SMDM 1B setting. BackPlay and the joint-training control use the same LBC data construction and inference policy, and differ only in whether $\theta$ is updated during correction-head training.}
    \label{tab:generative_fidelity}
    \centering
    \begin{tabular}{l@{\hspace{-0pt}}cc}
        \toprule
        \small
        Model & {Acc.(No Remask)} & {Acc.(Remask)} \\
        \midrule
        SFT Baseline ($\theta^*$) & $\mathbf{61.33}$ & -- \\
        Joint-Training Control $(\theta + \phi)^*$ & $60.56$ & $62.62$ \\
        BackPlay ($\theta^*, \phi^*$) & $\mathbf{61.33}$ & $\mathbf{63.46}$ \\
        \bottomrule
    \end{tabular}
\end{table}

Table~\ref{tab:generative_fidelity} reports the results. As expected, freezing $\theta^*$ preserves the original no-remask accuracy of the deployed generator (61.33\%), providing a sanity check that correction-head training does not alter the base model. In contrast, in our head-based SMDM 1B setup, jointly updating $\theta$ and $\phi$ slightly reduces no-remask accuracy to 60.56\%.

With self-correction enabled, both BackPlay and the joint-training control outperform the frozen SFT baseline without self-correction, while BackPlay achieves higher accuracy than the joint-training control (63.46\% vs.\ 62.62\%). This is consistent with the hypothesis that training the head on the fixed generator that is finally deployed yields more effective correction behavior in this setting.

\subsection{LLaDA 8B: End-to-End Benchmark Evaluation}
\label{sec:LLaDA}

To evaluate BackPlay at larger scale, we conduct experiments on LLaDA 8B \citep{nie2025large} across coding and mathematical reasoning tasks. For coding, we use LLaDA 8B Base as the base model and perform standard SFT on the first 500,000 samples from OpenCodeInstruct \citep{ahmad2025opencodeinstruct} to obtain $\theta^*$, then train the correction head $\phi^*$ on the same data. For math reasoning, we follow the same procedure using the first 500,000 samples from OpenMathInstruct \citep{toshniwal2024openmathinstruct}.

The correction head $\phi$ is a 3-layer Transformer that takes the output hidden state of the 31st block (out of 32) as input. Both the SFT baseline and BackPlay use the same semi-autoregressive generation protocol from LLaDA \citep{nie2025large}: block length 32, maximum generation length 1024, and early stopping upon detecting an end-of-sequence token. Unless otherwise stated, we keep the self-correction hyperparameters $(\tau, K, d, B)$ fixed across tasks and token-per-step settings. Full hyperparameters are provided in \Cref{sec:infer llada}.

\subsubsection{Coding Experiments}
\label{sec: llada coding}

We evaluate on the MBPP and HumanEval benchmarks. We observed that, across all methods, increasing the number of tokens generated per sampling step reduces $\text{Iter}_{\text{avg}}$ but also decreases accuracy, reflecting the standard accuracy--efficiency trade-off reported in LLaDA \citep{nie2025large}.

\begin{table}[htb!]
    \centering
    \caption{Coding results. BackPlay vs.\ SFT baseline on MBPP and HumanEval.}
    \label{tab:mbpp_results}\label{tab:humaneval_results}
    {\small
    \setlength{\tabcolsep}{3.5pt}
    \begin{tabular}{l l|cccc|cccc}
    \toprule
    Method & Metric 
    & \multicolumn{4}{c|}{MBPP} 
    & \multicolumn{4}{c}{HumanEval} \\
    
    & 
    & 1 & 2 & 3 & 4 
    & 1 & 2 & 3 & 4 \\
    \midrule
    
    \multirow{2}{*}{SFT} 
    & Acc(\%) 
    & \textit{39.8} & \textit{35.6} & \textit{30.4} & \textit{24.6} 
    & \textit{35.98} & \textit{28.66} & \textit{24.39} & \textit{20.73} \\
    
    & $\text{Iter}_{\text{avg}}$ 
    & \textit{119.4} & \textit{59.3} & \textit{44.3} & \textit{30.0} 
    & \textit{131.0} & \textit{67.1} & \textit{41.5} & \textit{29.2} \\
    
    \midrule
    
    \multirow{2}{*}{BackPlay} 
    & Acc(\%) 
    & \textit{40.2} & \textit{40.0} & \textit{35.8} & \textit{33.8} 
    & \textit{36.59} & \textit{33.54} & \textit{33.29} & \textit{27.44} \\
    
    & $\text{Iter}_{\text{avg}}$ 
    & \textit{134.0} & \textit{69.6} & \textit{45.6} & \textit{35.9} 
    & \textit{138.6} & \textit{71.1} & \textit{41.5} & \textit{35.5} \\
    
    \bottomrule
    \end{tabular}}
\end{table}

\looseness=-1 Table~\ref{tab:mbpp_results} summarizes the results. BackPlay consistently improves the accuracy--efficiency trade-off: at more aggressive multi-token decoding settings, BackPlay maintains higher accuracy than the SFT baseline at comparable or moderately larger decoding budgets. On MBPP, BackPlay with 2 tokens per step reaches 40.0\% accuracy, exceeding the baseline at 1 token per step (39.8\%) while reducing $\text{Iter}_{\text{avg}}$ from 119.4 to 69.6---a 41.7\% reduction. On HumanEval, BackPlay with 3 tokens per step attains 33.29\%, exceeding the baseline at 2 tokens per step (28.66\%) while reducing $\text{Iter}_{\text{avg}}$ from 67.1 to 41.5.

\begin{figure}[htb!]
    \centering
    \includegraphics[width=0.48\linewidth]{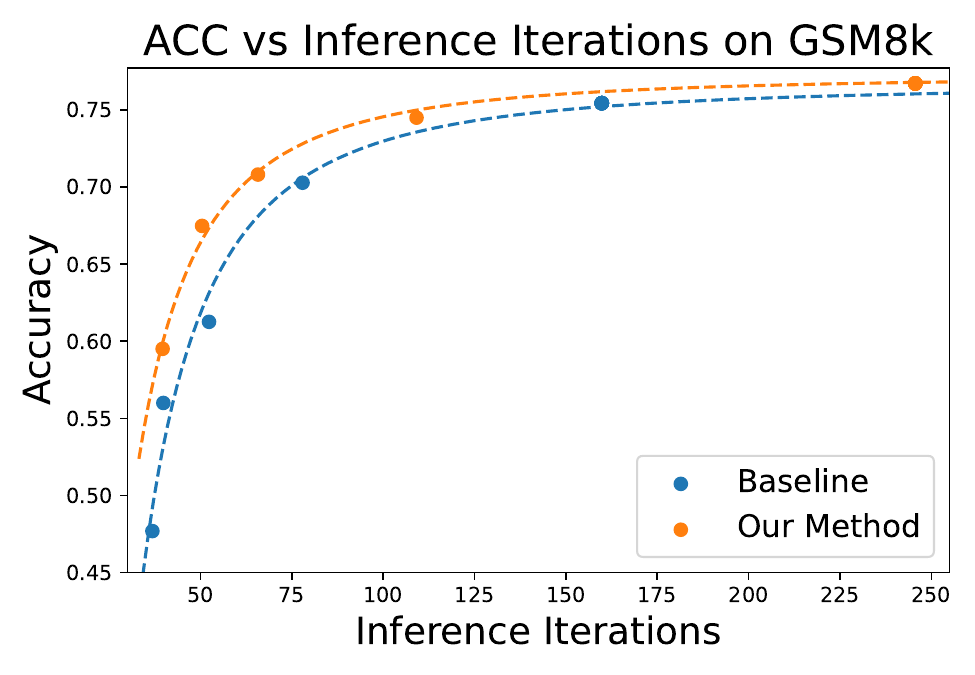}
    \includegraphics[width=0.48\linewidth]{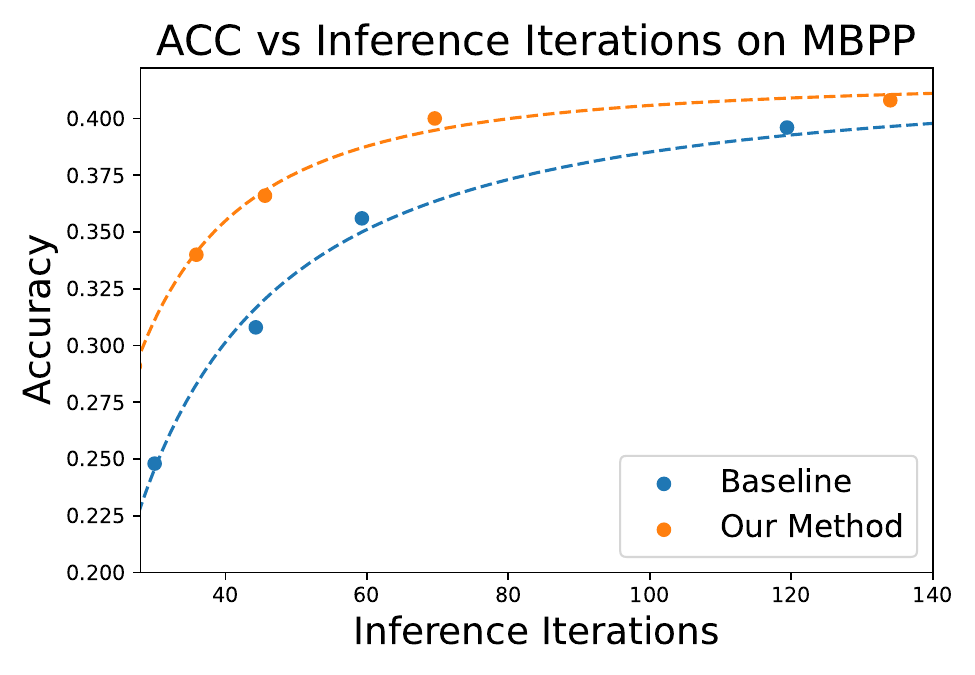}
    \caption{Plot of Accuracy vs.\ $\text{Iter}_{\text{avg}}$ for BackPlay and the SFT baseline on GSM8K and MBPP. Data from Tables~\ref{tab:mbpp_results} and~\ref{tab:gsm8k_results}. BackPlay improves the accuracy--efficiency trade-off under multi-token decoding.}
    \label{fig:baseline comparison}
\end{figure}

BackPlay is primarily intended for multi-token decoding regimes, where dependency errors are more prevalent. At the slowest setting (1 token per step), the corrector can still improve accuracy, but the additional intervention opportunities may increase $\text{Iter}_{\text{avg}}$. Figure~\ref{fig:baseline comparison} visualizes the trade-off relative to the SFT baseline on one representative coding benchmark (MBPP) and one representative reasoning benchmark (GSM8K); full numerical results for HumanEval and MATH are reported in Tables~\ref{tab:humaneval_results} and~\ref{tab:math_results}.

\subsubsection{Mathematical Reasoning Experiments}
\label{sec: llada math}

We evaluate zero-shot accuracy (Pass@1) on GSM8K \citep{cobbe2021training} and MATH \citep{hendrycks2021measuring}. Table~\ref{tab:gsm8k_results} shows that BackPlay consistently improves the accuracy--efficiency trade-off over the SFT baseline in the reported mathematical reasoning settings. On GSM8K, BackPlay with 3 tokens per step reaches 70.81\%, exceeding the baseline at 2 tokens per step (70.28\%) while reducing $\text{Iter}_{\text{avg}}$ from 77.9 to 65.7---a 15.7\% reduction. On MATH, BackPlay with 3 tokens per step reaches 33.6\%, exceeding the baseline at 2 tokens per step (31.6\%) while reducing $\text{Iter}_{\text{avg}}$ from 85.3 to 75.4. As in the coding experiments, the benefit is most pronounced under multi-token decoding.

\begin{table}[htb!]
    \centering
    \caption{Math reasoning results. BackPlay vs.\ SFT baseline on GSM8K and MATH.}
    \label{tab:gsm8k_results}\label{tab:math_results}
    {\small
    \setlength{\tabcolsep}{3pt}
    \begin{tabular}{l l|ccccc|cccc}
    \toprule
    Method & Metric 
    & \multicolumn{5}{c|}{GSM8K} 
    & \multicolumn{4}{c}{MATH} \\
    
    & 
    & 1 & 2 & 3 & 4 & 5 
    & 1 & 2 & 3 & 4 \\
    
    \midrule
    
    \multirow{2}{*}{SFT} 
    & Acc(\%) 
    & \textit{75.44} & \textit{70.28} & \textit{61.26} & \textit{56.18} & \textit{47.69} 
    & \textit{34.2} & \textit{31.6} & \textit{28.4} & \textit{22.4} \\
    
    & $\text{Iter}_{\text{avg}}$ 
    & \textit{159.8} & \textit{77.9} & \textit{52.3} & \textit{39.8} & \textit{36.8} 
    & \textit{190.5} & \textit{85.3} & \textit{61.1} & \textit{47.1} \\
    
    \midrule
    
    \multirow{2}{*}{BackPlay} 
    & Acc(\%) 
    & \textit{76.72} & \textit{74.30} & \textit{70.81} & \textit{67.48} & \textit{59.51} 
    & \textit{36.2} & \textit{34.0} & \textit{33.6} & \textit{28.4} \\
    
    & $\text{Iter}_{\text{avg}}$ 
    & \textit{245.6} & \textit{109.1} & \textit{65.7} & \textit{50.4} & \textit{39.6} 
    & \textit{324.42} & \textit{127.9} & \textit{75.4} & \textit{56.1} \\
    
    \bottomrule
    \end{tabular}}
\end{table}

\subsection{Controlled Ablation on Artifact Source}
\label{sec:ablation}

\looseness=-1 To evaluate the importance of realistic artifact construction, we conduct a controlled ablation: while maintaining the training recipe in \Cref{sec:DSC}, we replace the model-generated artifacts in LBC (\Cref{sec:FCA}) with a uniform distribution over the vocabulary, re-train the correction head, and re-evaluate under the same inference policy. This experiment is not a full reproduction of RemeDi; it is a controlled artifact-source ablation that keeps the backbone, head architecture, training recipe, and inference policy fixed and changes only the artifact source. We evaluate on MBPP and GSM8K under the LLaDA 8B setting described in \Cref{sec:LLaDA}.

\begin{table}[htb!]
    \centering
    \caption{Uniform-artifact control on MBPP and GSM8K.}
    \label{tab:mbpp_results_ablation}\label{tab:gsm8k_results_ablation}
    {\small
    \setlength{\tabcolsep}{3pt}
    \begin{tabular}{l l|cccc|ccccc}
    \toprule
    Method & Metric 
    & \multicolumn{4}{c|}{MBPP} 
    & \multicolumn{5}{c}{GSM8K} \\
    
    & 
    & 1 & 2 & 3 & 4 
    & 1 & 2 & 3 & 4 & 5 \\
    
    \midrule
    
    \multirow{2}{*}{Unif.\ Artifact} 
    & Acc(\%) 
    & \textit{40.0} & \textit{37.6} & \textit{33.8} & \textit{29.8} 
    & \textit{75.74} & \textit{70.74} & \textit{63.61} & \textit{58.38} & \textit{49.37} \\
    
    & $\text{Iter}_{\text{avg}}$ 
    & \textit{119.2} & \textit{60.4} & \textit{44.1} & \textit{29.6} 
    & \textit{158.1} & \textit{78.6} & \textit{52.7} & \textit{40.5} & \textit{36.4} \\
    
    \midrule
    
    \multirow{2}{*}{BackPlay} 
    & Acc(\%) 
    & \textit{40.2} & \textit{40.0} & \textit{35.8} & \textit{33.8} 
    & \textit{76.72} & \textit{74.30} & \textit{70.81} & \textit{67.48} & \textit{59.51} \\
    
    & $\text{Iter}_{\text{avg}}$ 
    & \textit{134.0} & \textit{69.6} & \textit{45.6} & \textit{35.9} 
    & \textit{245.6} & \textit{109.1} & \textit{65.7} & \textit{50.4} & \textit{39.6} \\
    
    \bottomrule
    \end{tabular}}
\end{table}

\begin{figure}[htb!]
    \centering
    \includegraphics[width=0.48\linewidth]{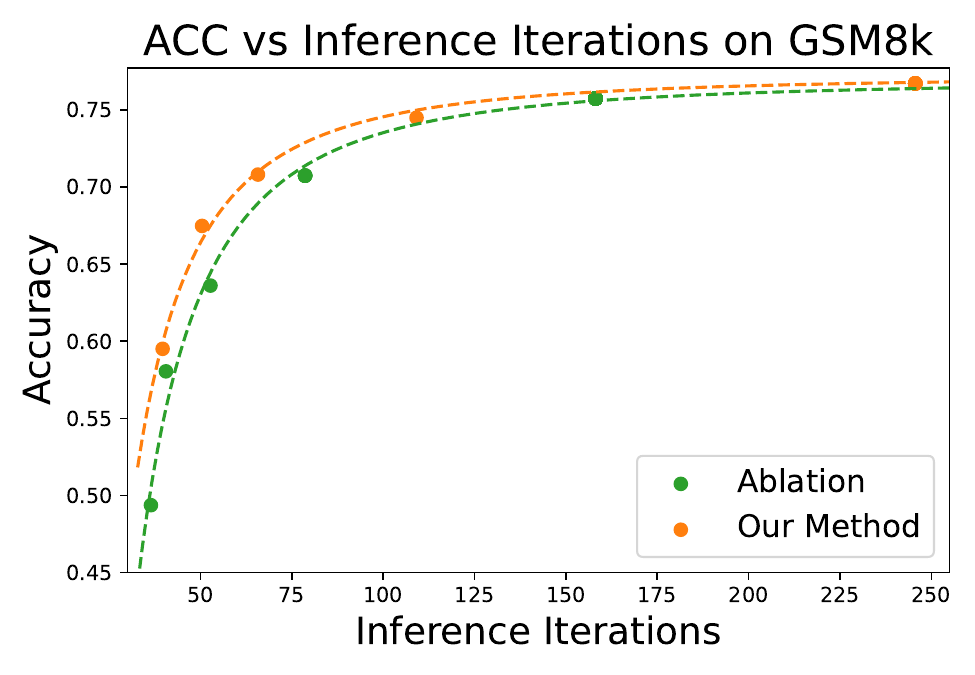}
    \includegraphics[width=0.48\linewidth]{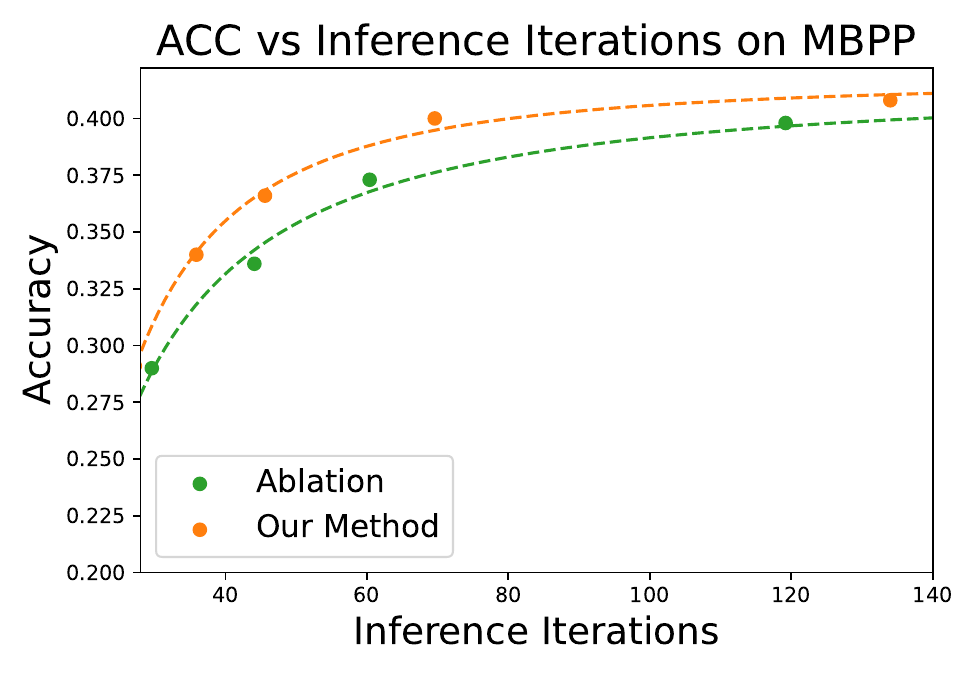}
    \caption{Accuracy vs.\ $\text{Iter}_{\text{avg}}$: BackPlay vs.\ the uniform-artifact control on GSM8K and MBPP. Replacing model-generated artifacts with uniform noise consistently shifts the curve downward. Data from Table~\ref{tab:mbpp_results_ablation}.}
    \label{fig:ablation comparison}
\end{figure}

Table~\ref{tab:mbpp_results_ablation} and Figure~\ref{fig:ablation comparison} show that replacing model-generated artifacts with uniform noise consistently shifts the accuracy--iteration curve downward across both benchmarks, supporting the importance of realistic, model-generated artifact construction for effective self-correction.

\subsection{Random-Remask Control}
\label{sec:random}

We compare BackPlay's learned remasking policy against a random-remask control. Specifically, we replace the error scores $e_i$ in the conservative adaptive remasking procedure (\Cref{alg:dsc_inference}) with independent random values sampled from $U[0,1]$, so that tokens are selected for remasking uniformly at random rather than by learned correctness scores. All other hyperparameters remain identical. Because our decoding protocol differs from \citet{wang2025remasking}, this experiment should be interpreted as a random-remask control within our inference framework rather than an exact reproduction of their full method.

\begin{table}
    \centering
    \caption{Random-remask control on GSM8K.}
    \label{tab:gsm8k_results_random}
    \setlength{\tabcolsep}{8pt}
    {\small
    \begin{tabular}{l l|ccccc}
    \toprule
    Method & Metric & 1 & 2 & 3 & 4 & 5 \\
    \midrule
    
    \multirow{2}{*}{Rand.} 
    & Acc(\%) 
    & 75.82 & 71.95 & 66.64 & 63.08 & 51.10 \\
    
    & $\text{Iter}_{\text{avg}}$ 
    & 338.4 & 155.6 & 75.3 & 58.7 & 40.9 \\
    
    \midrule
    
    \multirow{2}{*}{BackPlay} 
    & Acc(\%) 
    & 76.72 & 74.30 & 70.81 & 67.48 & 59.51 \\
    
    & $\text{Iter}_{\text{avg}}$ 
    & 245.6 & 109.1 & 65.7 & 50.4 & 39.6 \\
    
    \bottomrule
    \end{tabular}}
\end{table}

\begin{figure}
    \centering
    \includegraphics[width=0.5\textwidth]{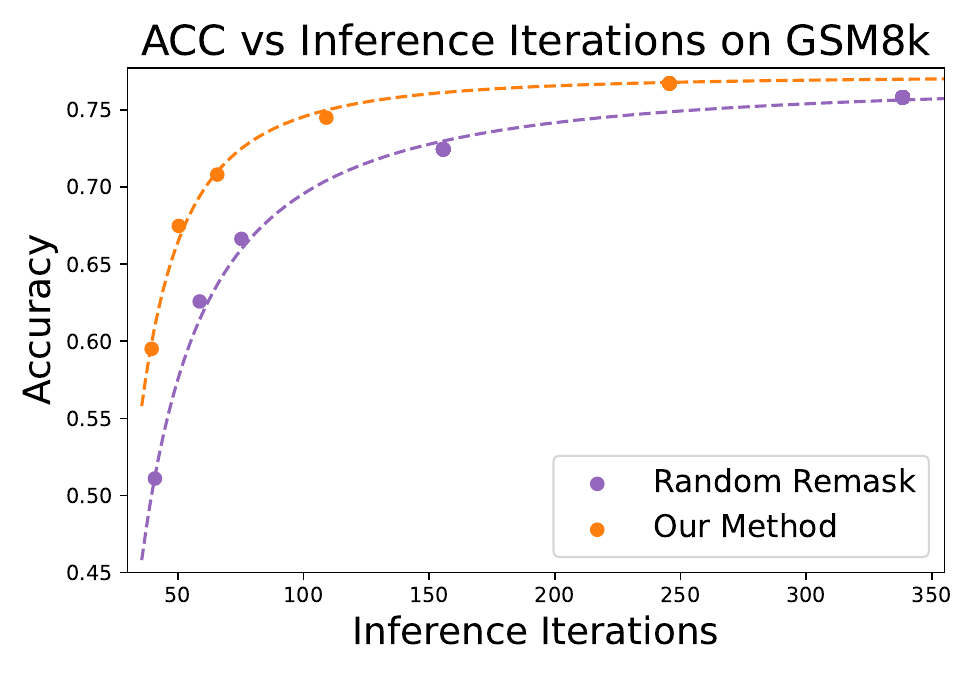}
    \vspace{-3pt}
    \caption{Accuracy vs.\ $\text{Iter}_{\text{avg}}$: BackPlay vs.\ random-remask on GSM8K. Data from Table~\ref{tab:gsm8k_results_random}.}
    \label{fig:random comparison}
\end{figure}

Although random remasking also improves accuracy over the SFT baseline, it cannot target erroneous tokens accurately, leading to unnecessary masking of correct tokens and redundant regeneration that increases $\text{Iter}_{\text{avg}}$. As shown in Table~\ref{tab:gsm8k_results_random} and Figure~\ref{fig:random comparison}, across all reported token-per-step settings on GSM8K, BackPlay achieves both higher accuracy and fewer DLM forward passes than the random-remask control. The gap indicates that the gain comes from targeted error selection rather than from simply remasking more tokens.

\section{Conclusion}

We propose BackPlay, a head-only self-correction framework for frozen finetuned Diffusion Language Models. BackPlay decouples corrector training from the generator: the finetuned DLM remains frozen, and only a lightweight correction head is trained on errors produced by the same fixed generator used at deployment, avoiding generator drift during head training. We introduce Look-back Correction, a training mechanism that injects predictions from earlier, more corrupted denoising states into later, richer contexts, enabling the head to leverage later context to detect earlier mistakes. At inference time, BackPlay converts the learned correctness scores into a conservative remasking policy that periodically revisits previously generated tokens. Experiments on mathematical reasoning and code generation benchmarks show that BackPlay improves the speed--quality trade-off of the underlying DLM under multi-token decoding.


\bibliography{colm2026_conference}

@article{nie2025large,
  title={Large language diffusion models},
  author={Nie, Shen and Zhu, Fengqi and You, Zebin and Zhang, Xiaolu and Ou, Jingyang and Hu, Jun and Zhou, Jun and Lin, Yankai and Wen, Ji-Rong and Li, Chongxuan},
  journal={arXiv preprint arXiv:2502.09992},
  year={2025}
}

@article{sahoo2024simple,
  title={Simple and effective masked diffusion language models},
  author={Sahoo, Subham Sekhar and Arriola, Marianne and Schiff, Yair and Gokaslan, Aaron and Marroquin, Edgar and Chiu, Justin T. and Rush, Alexander and Kuleshov, Volodymyr},
  journal={Advances in Neural Information Processing Systems},
  volume={37},
  pages={130136--130184},
  year={2024}
}

@article{nie2024scaling,
  title={Scaling up masked diffusion models on text},
  author={Nie, Shen and Zhu, Fengqi and Du, Chao and Pang, Tianyu and Liu, Qian and Zeng, Guangtao and Lin, Min and Li, Chongxuan},
  journal={arXiv preprint arXiv:2410.18514},
  year={2024}
}

@article{huang2025don,
  title={Don't Settle Too Early: Self-Reflective Remasking for Diffusion Language Models},
  author={Huang, Zemin and Wang, Yuhang and Chen, Zhiyang and Qi, Guo-Jun},
  journal={arXiv preprint arXiv:2509.23653},
  year={2025}
}

@article{kim2025fine,
  title={Fine-Tuning Masked Diffusion for Provable Self-Correction},
  author={Kim, Jaeyeon and Kim, Seunggeun and Lee, Taekyun and Pan, David Z and Kim, Hyeji and Kakade, Sham and Chen, Sitan},
  journal={arXiv preprint arXiv:2510.01384},
  year={2025}
}

@article{ahmad2025opencodeinstruct,
  title={{OpenCodeInstruct}: A Large-scale Instruction Tuning Dataset for Code {LLMs}},
  author={Ahmad, Wasi Uddin and Ficek, Aleksander and Samadi, Mehrzad and Huang, Jocelyn and Noroozi, Vahid and Majumdar, Somshubra and Ginsburg, Boris},
  journal={arXiv preprint arXiv:2504.04030},
  year={2025}
}

@article{toshniwal2024openmathinstruct,
  title={{OpenMathInstruct-2}: Accelerating {AI} for math with massive open-source instruction data},
  author={Toshniwal, Shubham and Du, Wei and Moshkov, Ivan and Kisacanin, Branislav and Ayrapetyan, Alexan and Gitman, Igor},
  journal={arXiv preprint arXiv:2410.01560},
  year={2024}
}

@article{wang2025remasking,
  title={Remasking discrete diffusion models with inference-time scaling},
  author={Wang, Guanghan and Schiff, Yair and Sahoo, Subham Sekhar and Kuleshov, Volodymyr},
  journal={arXiv preprint arXiv:2503.00307},
  year={2025}
}

@article{shi2024simplified,
  title={Simplified and generalized masked diffusion for discrete data},
  author={Shi, Jiaxin and Han, Kehang and Wang, Zhe and Doucet, Arnaud and Titsias, Michalis K.},
  journal={Advances in neural information processing systems},
  volume={37},
  pages={103131--103167},
  year={2024}
}

@article{gat2024discrete,
  title={Discrete flow matching},
  author={Gat, Itai and Remez, Tal and Shaul, Neta and Kreuk, Felix and Chen, Ricky TQ and Synnaeve, Gabriel and Adi, Yossi and Lipman, Yaron},
  journal={Advances in Neural Information Processing Systems},
  volume={37},
  pages={133345--133385},
  year={2024}
}

@article{ye2025dream,
  title={Dream {7B}: Diffusion large language models},
  author={Ye, Jiacheng and Xie, Zhihui and Zheng, Lin and Gao, Jiahui and Wu, Zirui and Jiang, Xin and Li, Zhenguo and Kong, Lingpeng},
  journal={arXiv preprint arXiv:2508.15487},
  year={2025}
}

@article{liu2024discrete,
  title={Discrete copula diffusion},
  author={Liu, Anji and Broadrick, Oliver and Niepert, Mathias and Broeck, Guy Van den},
  journal={arXiv preprint arXiv:2410.01949},
  year={2024}
}

@article{xu2024energy,
  title={Energy-based diffusion language models for text generation},
  author={Xu, Minkai and Geffner, Tomas and Kreis, Karsten and Nie, Weili and Xu, Yilun and Leskovec, Jure and Ermon, Stefano and Vahdat, Arash},
  journal={arXiv preprint arXiv:2410.21357},
  year={2024}
}

@article{kang2025parallelbench,
  title={{ParallelBench}: Understanding the trade-offs of parallel decoding in diffusion {LLMs}},
  author={Kang, Wonjun and Galim, Kevin and Oh, Seunghyuk and Lee, Minjae and Zeng, Yuchen and Zhang, Shuibai and Hooper, Coleman and Hu, Yuezhou and Koo, Hyung Il and Cho, Nam Ik and others},
  journal={arXiv preprint arXiv:2510.04767},
  year={2025}
}

@article{zhao2024informed,
  title={Informed correctors for discrete diffusion models},
  author={Zhao, Yixiu and Shi, Jiaxin and Chen, Feng and Druckmann, Shaul and Mackey, Lester and Linderman, Scott},
  journal={arXiv preprint arXiv:2407.21243},
  year={2024}
}

@article{peng2025path,
  title={Path planning for masked diffusion model sampling},
  author={Peng, Fred Zhangzhi and Bezemek, Zachary and Patel, Sawan and Rector-Brooks, Jarrid and Yao, Sherwood and Bose, Avishek Joey and Tong, Alexander and Chatterjee, Pranam},
  journal={arXiv preprint arXiv:2502.03540},
  year={2025}
}

@article{deng2023implicit,
  title={Implicit chain of thought reasoning via knowledge distillation},
  author={Deng, Yuntian and Prasad, Kiran and Fernandez, Roland and Smolensky, Paul and Chaudhary, Vishrav and Shieber, Stuart},
  journal={arXiv preprint arXiv:2311.01460},
  year={2023}
}

@article{cobbe2021training,
  title={Training verifiers to solve math word problems},
  author={Cobbe, Karl and Kosaraju, Vineet and Bavarian, Mohammad and Chen, Mark and Jun, Heewoo and Kaiser, Lukasz and Plappert, Matthias and Tworek, Jerry and Hilton, Jacob and Nakano, Reiichiro and others},
  journal={arXiv preprint arXiv:2110.14168},
  year={2021}
}

@article{hendrycks2021measuring,
  title={Measuring mathematical problem solving with the {MATH} dataset},
  author={Hendrycks, Dan and Burns, Collin and Kadavath, Saurav and Arora, Akul and Basart, Steven and Tang, Eric and Song, Dawn and Steinhardt, Jacob},
  journal={arXiv preprint arXiv:2103.03874},
  year={2021}
}

@article{lou2023discrete,
  title={Discrete diffusion modeling by estimating the ratios of the data distribution},
  author={Lou, Aaron and Meng, Chenlin and Ermon, Stefano},
  journal={arXiv preprint arXiv:2310.16834},
  year={2023}
}

@article{austin2021structured,
  title={Structured denoising diffusion models in discrete state-spaces},
  author={Austin, Jacob and Johnson, Daniel D. and Ho, Jonathan and Tarlow, Daniel and van den Berg, Rianne},
  journal={Advances in neural information processing systems},
  volume={34},
  pages={17981--17993},
  year={2021}
}

@inproceedings{he2023diffusionbert,
  title={{DiffusionBERT}: Improving generative masked language models with diffusion models},
  author={He, Zhengfu and Sun, Tianxiang and Tang, Qiong and Wang, Kuanning and Huang, Xuanjing and Qiu, Xipeng},
  booktitle={Proceedings of the 61st annual meeting of the association for computational linguistics (volume 1: Long papers)},
  pages={4521--4534},
  year={2023}
}

@article{touvron2023llama,
  title={Llama 2: Open foundation and fine-tuned chat models},
  author={Touvron, Hugo and Martin, Louis and Stone, Kevin and Albert, Peter and Almahairi, Amjad and Babaei, Yasmine and Bashlykov, Nikolay and Batra, Soumya and Bhargava, Prajjwal and Bhosale, Shruti and others},
  journal={arXiv preprint arXiv:2307.09288},
  year={2023}
}

@article{zhu2025llada,
  title={{LLaDA} 1.5: Variance-Reduced Preference Optimization for Large Language Diffusion Models},
  author={Zhu, Fengqi and Wang, Rongzhen and Nie, Shen and Zhang, Xiaolu and Wu, Chunwei and Hu, Jun and Zhou, Jun and Chen, Jianfei and Lin, Yankai and Wen, Ji-Rong and others},
  journal={arXiv preprint arXiv:2505.19223},
  year={2025}
}
\bibliographystyle{colm2026_conference}

\newpage
\appendix
\section{Implementation Details of LBC}

We summarize the Look-back Correction algorithm in \Cref{alg:lbc_training}.
\begin{algorithm}
\caption{Look-back Correction (LBC) Training Recipe}
\label{alg:lbc_training}
\begin{algorithmic}[1]
\STATE \textbf{Input:} Clean data $\mathbf{x}_0$, frozen DLM parameters $\theta^*$, step size $\Delta t$, sequence length $L$
\STATE Sample time step $t \sim \mathcal{U}[0, 1{-}\Delta t]$
\STATE Sample state $\mathbf{x}_t \sim q_{t|0}(\cdot|\mathbf{x}_0)$
\STATE Sample forward interval $t' \sim \mathcal{U}[\Delta t, 1{-}t]$
\STATE Sample more corrupted state $\mathbf{x}_{t+t'} \sim q_{t+t'|0, t}(\cdot|\mathbf{x}_0, \mathbf{x}_{t})$
\STATE Generate artifact sequence: $\mathbf{y} \sim p_{\theta^*}(\mathbf{x}_{0} | \mathbf{x}_{t+t'})$
\STATE Identify newly unmasked positions: $\mathcal{I}_{\text{new}} = \{i : x_{t+t'}^i = \text{[MASK]} \;\text{and}\; x_t^i \neq \text{[MASK]}\}$
\STATE Calculate subset size: $k = \min(\lceil L \cdot \Delta t \rceil,\, |\mathcal{I}_{\text{new}}|)$
\STATE Select top-$k$ confident indices from $\mathcal{I}_{\text{new}}$: \\
$\mathcal{M} = \text{arg-top-}k_{i \in \mathcal{I}_{\text{new}}} \left(\max_v\, p_{\theta^*}(v \mid \mathbf{x}_{t+t'}) \right)$
\STATE Construct training sequence: $\mathbf{z}_t = \text{replace}(\mathbf{x}_t, \mathbf{y}, \mathcal{M})$
\STATE \textbf{Return:} Training pair $(\mathbf{z}_t, \mathbf{x}_0)$
\end{algorithmic}
\end{algorithm}

\section{Training Details on SMDM 1B}

For the SFT stage, we use the 1B pretrained model from SMDM \citep{nie2024scaling} as the base model and train on the augmented data from \citet{deng2023implicit}. Following \citet{nie2024scaling}, we initially train for 40 epochs, then extend to 80 epochs after observing continued accuracy gains on GSM8K. The training configuration is listed in \Cref{tab:smdm sft}.

\begin{table}[htb!]
\centering
\caption{SFT configuration for SMDM 1B.}
\label{tab:smdm sft}
\begin{tabular}{lc}
\toprule

\multicolumn{2}{l}{\textbf{Base Model}} \\
Architecture      & SMDM \citep{nie2024scaling} \\
Parameters        & 1028M \\
Tokenizer         & LLaMA2 Tokenizer \citep{touvron2023llama} \\
\midrule

\multicolumn{2}{l}{\textbf{Training}} \\
Sequence Length   & 256 \\
Optimizer            & AdamW \\
Learning Rate        & $2 \times 10^{-4}$ \\
Weight Decay         & 0.1 \\
Global Batch Size    & 256 \\
$\beta_1$                  & 0.9 \\
$\beta_2$                & 0.95 \\
Warmup Ratio        & 0.01 \\
Learning Rate Schedule  & cosine\\

\bottomrule
\end{tabular}
\end{table}

The correction head is trained on the same augmented data for 20 epochs. The training configuration is summarized in \Cref{tab:smdm_head}. For the joint-training control, we use the same hyperparameters as SFT (\Cref{tab:smdm sft}) and tune $\gamma \in \{0.01, 0.1, 0.5\}$, reporting the best result.

\begin{table}[htb!]
\centering
\begin{minipage}[t]{0.48\linewidth}
\centering
\caption{Correction head training config for SMDM 1B.}
\label{tab:smdm_head}
\begin{tabular}{lc}
\toprule
\multicolumn{2}{l}{\textbf{Model}} \\
Architecture      & 2-layer Transformer \\
Parameters        & 90M \\
\midrule
\multicolumn{2}{l}{\textbf{Training}} \\
Step Size $\Delta t$ in LBC & $\frac{1}{32}$\\
Sequence Length   & 256 \\
Optimizer            & AdamW \\
Learning Rate        & $2 \times 10^{-3}$ \\
Weight Decay         & 0.1 \\
Global Batch Size    & 256 \\
$\beta_1$            & 0.9 \\
$\beta_2$            & 0.95 \\
Warmup Ratio         & 0.01 \\
Learning Rate Schedule & cosine \\
\bottomrule
\end{tabular}
\end{minipage}
\hfill
\begin{minipage}[t]{0.48\linewidth}
\centering
\caption{Inference config for SMDM 1B.}
\label{tab:smdm_infer}
\begin{tabular}{lc}
\toprule
\multicolumn{2}{l}{\textbf{Inference}} \\
Generation Length   & 256 \\
Tokens per step $k$ & 2 \\
\hdashline
$\,$ \\
\multicolumn{2}{l}{\textbf{Conservative Adaptive Remasking}} \\
Error threshold $\tau$ & 0.75 \\
Remasking budget $K$ & 2 \\
Correction stride $d$ & 4 \\
Block buffer $B$ & 4 \\
\bottomrule
\end{tabular}
\end{minipage}
\end{table}

\section{Inference Details on SMDM 1B}
\label{sec:infer smdm}

For inference without remasking, following the setting in SMDM and LLaDA \citep{nie2025large}, we use the confidence-based strategy shown in \Cref{alg:confidence_inference}: at each iteration, the top-$k$ masked positions with the highest predicted confidence are demasked. For inference with remasking, both BackPlay and the joint-training control use the conservative adaptive remasking procedure in \Cref{alg:dsc_inference}. Hyperparameters are summarized in \Cref{tab:smdm_infer}.

\begin{algorithm}
\caption{Confidence-based inference strategy from LLaDA \citep{nie2025large}}
\label{alg:confidence_inference}
\begin{algorithmic}[1]

\REQUIRE
Generative model $\theta^*$,
tokens per step $k$,
generation length $L$.

\STATE $\Delta t \leftarrow \frac{k}{L}$, \quad $\mathbf{x}_1 \leftarrow [\text{MASK}]^L$, \quad $t \leftarrow 1$, \quad $N \leftarrow 0$

\WHILE{$t > 0$}
    \STATE $t_{\text{new}} \leftarrow \max \{t - \Delta t, 0\}, \,\, x_{t_\text{new}} \leftarrow x_t, \,\, \mathcal{M} \leftarrow \{ i \mid x_t^{(i)} = [\text{MASK}] \}$
    \STATE $P_{\text{demask}} \leftarrow p_{\theta^*}(\mathbf{x}_{0} \mid \mathbf{x}_t)$
    \STATE $\mathcal{I} \leftarrow \text{arg-top-}k_{i \in \mathcal{M}} \left(P_{\text{demask}}^{(i)} \right)$

    \STATE Sample $\mathbf{x}_{t_{\text{new}}}[i] \sim P_{\text{demask}}[i]$ for all $i \in \mathcal{I}$

    \STATE $t \leftarrow t_{\text{new}}$, \quad $N \leftarrow N + 1$
\ENDWHILE

\end{algorithmic}
\end{algorithm}

\section{Training Details on LLaDA 8B}

\subsection{Supervised Finetuning (SFT)}
\label{sec:appendix sft}

Our SFT procedure for LLaDA 8B Base follows \citet{nie2025large}. We set the maximum sequence length to 1024 and pad all sequences to this length using the end-of-sequence token [EoS].

In LLaDA \citep{nie2025large}, padding [EoS] tokens are treated as normal tokens and randomly masked with the same probability as response tokens. However, when most training sequences are much shorter than the maximum length, the abundance of [EoS] tokens causes the model to predict [EoS] with high probability, leading to premature termination of generation. This issue is also noted in the LLaDA codebase.

To address this, we apply a simple modification during SFT: the first 16 [EoS] tokens after the response are treated as normal tokens and may be masked, while all remaining [EoS] padding tokens are kept unmasked. This effectively limits the number of [EoS] tokens the model must predict and resolves the early-termination issue.

We use the standard demasking loss (\Cref{eq:sft loss}) and train LLaDA 8B Base on the first 500,000 samples from OpenCodeInstruct \citep{ahmad2025opencodeinstruct} for coding experiments (\Cref{sec: llada coding}) and the first 500,000 samples from OpenMathInstruct \citep{toshniwal2024openmathinstruct} for math reasoning experiments (\Cref{sec: llada math}). Hyperparameters are summarized in \Cref{tab:llada sft}.

\begin{table}[htb!]
\centering
\caption{SFT configuration for LLaDA 8B.}
\label{tab:llada sft}
\begin{tabular}{lc}
\toprule

\multicolumn{2}{l}{\textbf{Base Model}} \\
Architecture      & LLaDA Base \citep{nie2025large} \\
Parameters        & 8B \\
Tokenizer         & LLaDA Tokenizer \citep{nie2025large} \\
\midrule

\multicolumn{2}{l}{\textbf{Training}} \\
Sequence Length   & 1024 \\
Optimizer            & AdamW \\
Learning Rate        & $2.5 \times 10^{-5}$ \\
Weight Decay         & 0.1 \\
Global Batch Size    & 256 \\
$\beta_1$                  & 0.9 \\
$\beta_2$                & 0.95 \\
Warmup Ratio        & 0.01 \\
Learning Rate Schedule  & cosine\\

\bottomrule
\end{tabular}
\end{table}

\subsection{Training of the correction head}
\label{sec:llada head hyper}

The correction head is trained after the SFT stage, with the DLM parameters $\theta^{*}$ frozen. We use a 3-layer Transformer block as the correction head $\phi$, taking the output of the 31st layer of the DLM (out of 32 layers) as input. The head outputs a correctness score $c_i \in [0,1]$ for each visible position, estimating $P(z_t^i = x_0^i \mid \mathbf{z}_t)$. The correction head has approximately 600M parameters, substantially smaller than the 8B base model. The training procedure follows \Cref{sec:DSC} and \Cref{sec:FCA}, using the same dataset as for $\theta^{*}$ (\Cref{sec:appendix sft}). Consistent with the SFT stage, the first 16 [EoS] tokens are treated as normal tokens, and remaining [EoS] padding tokens are kept unmasked during the construction of $\mathbf{z}_t$ (\Cref{eq: Dsc loss error}). Hyperparameters are summarized in \Cref{tab:llada_head}.

\subsection{Training details of the ablation study}

For the artifact-source ablation in \Cref{sec:ablation}, we retrain only the correction head on the modified training data (uniform artifacts instead of model-generated artifacts). All training and inference hyperparameters remain identical to those in \Cref{sec:llada head hyper}.

\begin{table}[htb!]
\centering
\begin{minipage}[t]{0.48\linewidth}
\centering
\caption{Correction head training config for LLaDA 8B.}
\label{tab:llada_head}
\begin{tabular}{lc}
\toprule
\multicolumn{2}{l}{\textbf{Model}} \\
Architecture      & 3-layer Transformer \\
Parameters        & ${\sim}$600M \\
\midrule
\multicolumn{2}{l}{\textbf{Training}} \\
Step Size $\Delta t$ in LBC & $\frac{1}{64}$\\
Sequence Length   & 1024 \\
Optimizer            & AdamW \\
Learning Rate        & $5 \times 10^{-4}$ \\
Weight Decay         & 0.1 \\
Global Batch Size    & 256 \\
$\beta_1$            & 0.9 \\
$\beta_2$            & 0.95 \\
Warmup Ratio         & 0.01 \\
Learning Rate Schedule & cosine \\
\bottomrule
\end{tabular}
\end{minipage}
\hfill
\begin{minipage}[t]{0.48\linewidth}
\centering
\caption{Inference config for LLaDA 8B.}
\label{tab:llada_infer}
\begin{tabular}{lc}
\toprule
\multicolumn{2}{l}{\textbf{Inference}} \\
Generation Length   & 1024 \\
Tokens per step $k$ & -- \\

\hdashline

\multicolumn{2}{l}{\textbf{Conservative Adaptive Remasking}} \\
Error threshold $\tau$ & 0.75 \\
Remasking budget $K$ & 2 \\
Correction stride $d$ & 4 \\
Block buffer $B$ & 4 \\
\bottomrule
\end{tabular}
\end{minipage}
\end{table}

\section{Inference Details on LLaDA 8B}
\label{sec:infer llada}

For the SFT baseline, we use the confidence-based strategy from LLaDA (\Cref{alg:confidence_inference}). For BackPlay, we use the conservative adaptive remasking procedure in \Cref{alg:dsc_inference}. Tokens per step $k$ is varied across experiments to compare the accuracy--efficiency trade-off, as discussed in Section~4 and reported in the result tables. All other inference hyperparameters are summarized in \Cref{tab:llada_infer}. For the random-remask control (\Cref{sec:random}), all hyperparameters are kept identical to those of BackPlay, except that the error scores $e_i$ are replaced with independent random values sampled from $U[0, 1]$.

\subsection{Sensitivity to remasking threshold $\tau$}

To examine the sensitivity of the error threshold $\tau$, we run an additional inference experiment with $\tau = 0.6$ while keeping all other hyperparameters identical to \Cref{tab:llada_infer}, and evaluate on MBPP. As shown in \Cref{tab:mbpp_results_tau}, lowering $\tau$ slightly increases accuracy at larger tokens-per-step settings but also increases $\text{Iter}_{\text{avg}}$, reflecting a standard precision--recall trade-off in error detection rather than brittle behavior.

\begin{table}[htb!]
    \centering
    \setlength{\tabcolsep}{3pt}
    \caption{Sensitivity of $\tau$ on MBPP. Lowering $\tau$ from 0.75 to 0.6 yields a modest accuracy--iteration trade-off.}
    \label{tab:mbpp_results_tau}
    {\small
    \begin{tabular}{c|c|c|c}
    \toprule
    Method & Tokens/Step & Acc(\%) & $\text{Iter}_{\text{avg}}$ \\
    \midrule
    \multirow{4}{*}{$\tau=0.6$} & 1 & \textit{40.2} & \textit{138.2} \\
    & 2 & \textit{40.0} & \textit{71.6} \\
    & 3 & \textit{36.4} & \textit{48.2} \\
    & 4 & \textit{34.2} & \textit{36.4} \\
    \midrule
    \multirow{4}{*}{$\tau = 0.75$} & 1 & \textit{40.2} & \textit{134.0} \\
    & 2 & \textit{40.0} & \textit{69.6} \\
    & 3 & \textit{35.8} & \textit{45.6} \\
    & 4 & \textit{33.8} & \textit{35.9} \\
    \bottomrule
    \end{tabular}}

\end{table}

\subsection{Sensitivity to Block buffer $B$}

To examine the sensitivity of the Block buffer $B$, we run an additional inference experiment with $B = 6$ while keeping all other hyperparameters identical to \Cref{tab:llada_infer}, and evaluate on MBPP. As shown in \Cref{tab:mbpp_results_B}, setting $B = 6$ only has a very slight impact on the final result.

\begin{table}[htb!]
    \centering
    \setlength{\tabcolsep}{3pt}
    \caption{Sensitivity of $B$ on MBPP. Increasing $B$ from 4 to 6 only has a very slight impact.}
    \label{tab:mbpp_results_B}
    {\small
    \begin{tabular}{c|c|c|c}
    \toprule
    Method & Tokens/Step & Acc(\%) & $\text{Iter}_{\text{avg}}$ \\
    \midrule
    \multirow{4}{*}{$B=6$} & 1 & \textit{40.2} & \textit{132.4} \\
    & 2 & \textit{39.8} & \textit{68.8} \\
    & 3 & \textit{35.8} & \textit{45.4} \\
    & 4 & \textit{33.8} & \textit{35.8} \\
    \midrule
    \multirow{4}{*}{$B = 4$} & 1 & \textit{40.2} & \textit{134.0} \\
    & 2 & \textit{40.0} & \textit{69.6} \\
    & 3 & \textit{35.8} & \textit{45.6} \\
    & 4 & \textit{33.8} & \textit{35.9} \\
    \bottomrule
    \end{tabular}}

\end{table}

\end{document}